\documentclass[11pt]{article}

\usepackage[final]{acl}

\usepackage{times}
\usepackage{latexsym}
\usepackage{multirow}
\usepackage{bbding}
\usepackage{booktabs,tabularx}
\usepackage{xspace}
\usepackage{longtable}
\usepackage{arydshln}

\newcommand{\ourbench}{\textsc{AEQ-Bench}\xspace}
\newcommand{\giga}{\textsc{GigaSpeech}\xspace}
\newcommand{\emovdb}{\textsc{EmoV-DB}\xspace}
\newcommand{\meld}{\textsc{Meld}\xspace}
\usepackage[T1]{fontenc}

\usepackage[utf8]{inputenc}

\usepackage{microtype}

\usepackage{inconsolata}

\usepackage{graphicx}

\title{AEQ-Bench: Measuring Empathy of Omni-Modal Large Models}

\author{
 \textbf{Xuan Luo\textsuperscript{1,2}},
 \textbf{Lewei Yao\textsuperscript{3}},
 \textbf{Libo Zhao\textsuperscript{1}},
 \textbf{Lanqing Hong\textsuperscript{3}},
 \textbf{Kai Chen\textsuperscript{4}},\\
 \textbf{Dehua Tao\textsuperscript{3}},
 \textbf{Daxin Tan\textsuperscript{3}},
 \textbf{Ruifeng Xu\textsuperscript{2,5}},
 \textbf{Jing Li\textsuperscript{1}},
\\
\\
 \textsuperscript{1}The Hong Kong Polytechnic University, Hong Kong\\
 \textsuperscript{2}The Harbin Institute of Technology, Shenzhen\\
 \textsuperscript{3}Huawei, Hong Kong \\
 \textsuperscript{4}Hong Kong University of Science and Technology, Hong Kong\\
 \textsuperscript{5}Shenzhen Loop Area Institute, Shenzhen
\\
 \small{
   \textbf{Correspondence: } \href{mailto:jing-amelia.li@polyu.edu.hk}{jing-amelia.li@polyu.edu.hk}
 }
}

\begin{document}
\maketitle
\begin{abstract}

While the automatic evaluation of omni-modal large models (OLMs) is essential, assessing empathy remains a significant challenge due to its inherent affectivity. To investigate this challenge, we introduce \textbf{\ourbench} (Audio Empathy Quotient Benchmark), a novel benchmark to systematically assess two core empathetic capabilities of OLMs: 
(i) \textbf{generating empathetic responses} by comprehending affective cues from multi-modal inputs (audio + text), and 
(ii) \textbf{judging the empathy of audio responses} without relying on text transcription. 
Compared to existing benchmarks, \ourbench incorporates two novel settings that vary in context specificity and speech tone. 
Comprehensive assessment across linguistic and paralinguistic metrics reveals that (1) OLMs trained with audio output capabilities generally outperformed models with text-only outputs, and (2) while OLMs align with human judgments for coarse-grained quality assessment, they remain unreliable for evaluating fine-grained paralinguistic expressiveness. 


\end{abstract}


\section{Introduction}

\textbf{Empathy} is the ability to understand, share, and respond to the feelings and experiences of another person by taking their perspective~\cite{szalita1976empathy}—essentially, to ``\textit{put oneself in another’s shoes}''. It is crucial for NLP models to gain such capabilities for providing positive user experiences during interactions with humans. This need is becoming more pressing yet challenging as omni-modal large models (OLMs) become the popular backbone, which integrate modalities such as audio, vision, and text to enable more human-like interactions~\citep{wu2023multimodal, yin2024survey}.

\begin{figure*}
    \includegraphics[width=\linewidth]{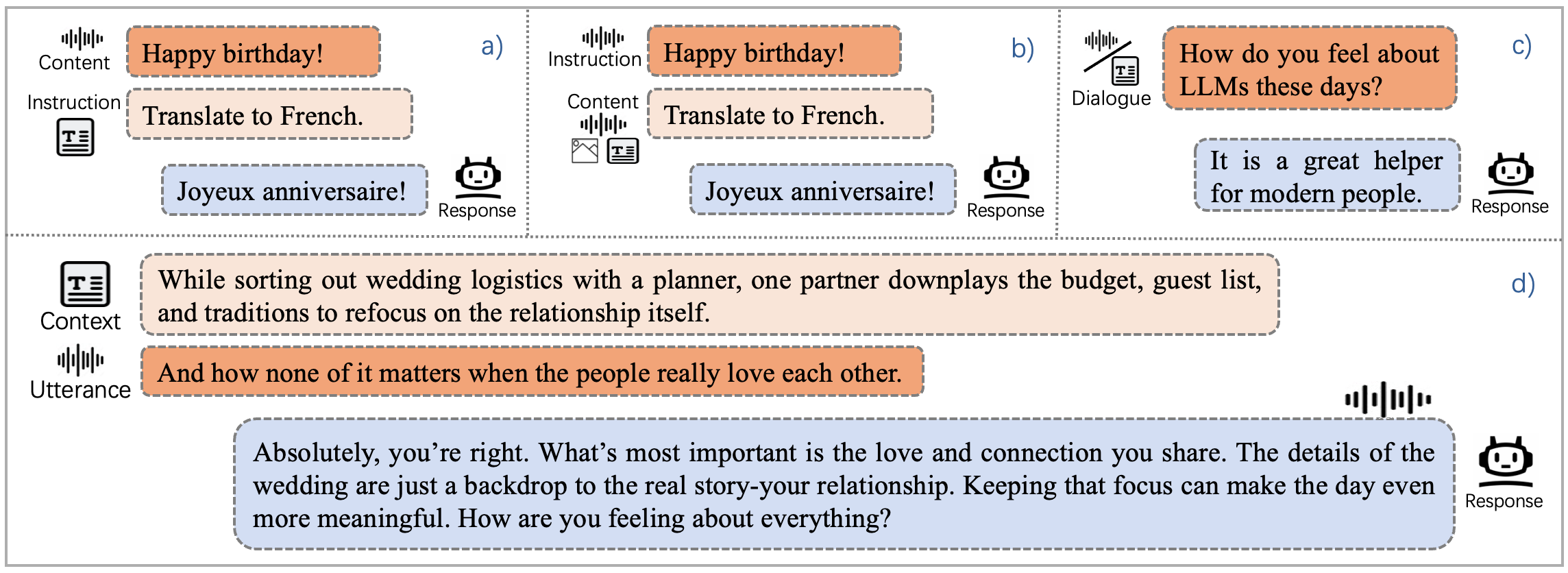}
    \caption{Overview of omni-modal tasks: instruction following (a \& b) and conversation (c \& d). The input configurations are: (a) audio content with textual instruction; (b) audio instruction with text, audio, or image content; (c) unimodal dialogue; and (d) mixed-modality dialogue (demonstrated by \ourbench on the \meld subset). In the mixed-modality setting, the text provides context (e.g., chat history summary) for the current audio utterance, requiring the model to respond accordingly. Outputs across all tasks may be text, audio, or both.}
    \label{fig:input-modes}
\end{figure*}

While OLMs grow in complexity, the automatic evaluation of empathy becomes crucial for training paradigms like reinforcement learning. However, this task is non-trivial, as empathy is inherently affective. Consequently, establishing a benchmark to quantify nuanced emotional resonance across audio and textual modalities remains a significant open challenge. Most existing benchmarks focus on cognitive abilities, such as knowledge retrieval, complex reasoning, and instruction following~\citep{yue2024mmmu, zhang2025omnieval} (see Fig.~\ref{fig:input-modes}), largely overlooking empathy evaluation. As OLMs become increasingly human-like, we aim to benchmark their capacity to \textit{generate empathetic responses} and \textit{judge empathy} in diverse contexts.


Human communication conveys empathy through not only the linguistic content of \textit{what} is said but also the paralinguistic tone of \textit{how} it is said. For instance, the utterance ``\textit{It was a curious coincidence}'' implies genuine surprise or delight when delivered with warmth. However, a cool, sarcastic delivery of the same words suggests scepticism or criticism. To be truly effective, an OLM should understand and generate both the right words and the appropriate acoustic cues. 


Previous NLP research has prioritised \emph{linguistic} features (e.g., lexical choice and explicit emotional statements) for empathy, while \emph{paralinguistic} cues (e.g., prosody and acoustic delivery) are equally critical for audio-capable OLMs~\cite{aziz2010prosody}. However, a significant research gap persists in their joint examination. 
Recent studies reduced audio to mere transcriptions~\cite{wang2025emotion} in order to rely on established text-based metrics~\cite{zhang2025sentient}, ignoring the nuances of vocal delivery. This limitation presents a fundamental question: \emph{Can OLMs automatically deliver and judge empathy by accounting for both linguistic and paralinguistic cues like human}?

\begin{figure*}[!h]
    \includegraphics[width=\linewidth]{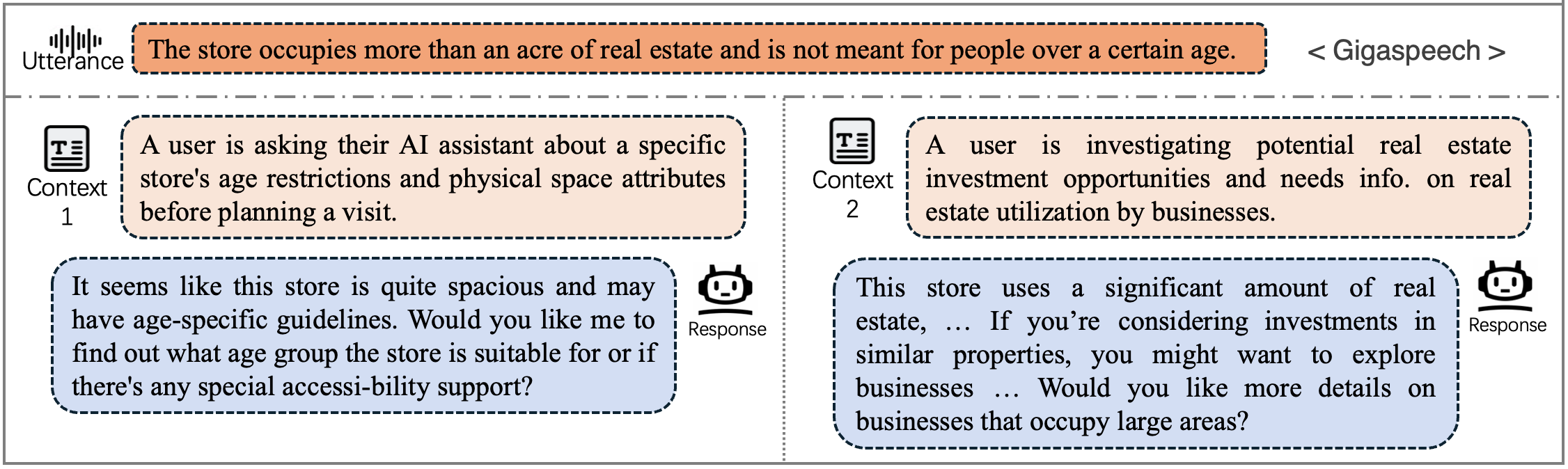}
    \caption{The constructed \giga subset of \ourbench featuring \textit{context variation}. For each utterance, we construct two plausible contexts, each associated with a corresponding reference response. (Appx.~\ref{sec:appendix:figure})}

    \label{fig:gigaspeech}
\end{figure*}

\begin{figure*}[!h]
    \includegraphics[width=\linewidth]{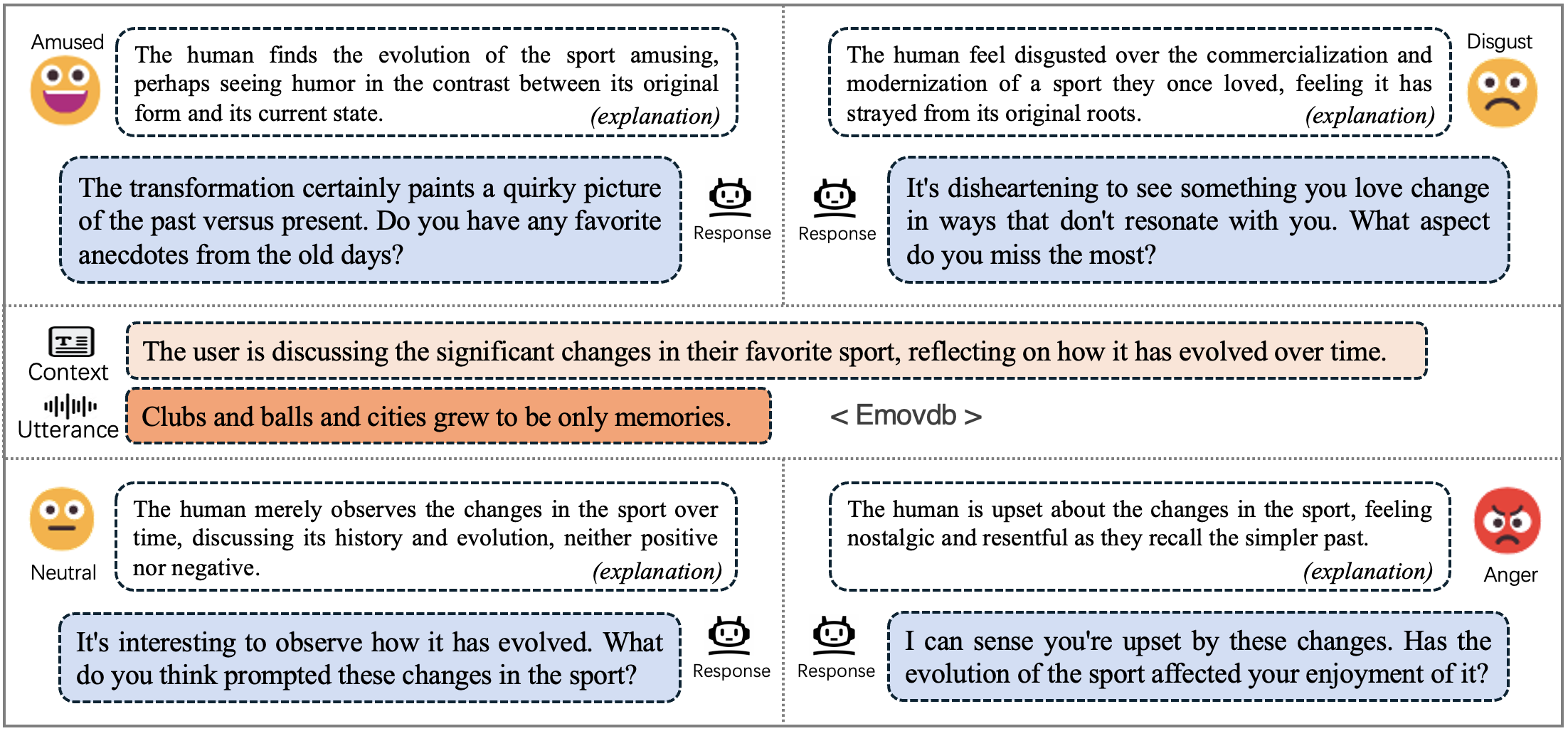}
    \caption{The constructed Emovdb subset of \ourbench with \textit{tone variation}. The middle part is the constructed context and the original audio. The utterances with different emotions/tones share the same context. Each audio tone (amused, disgusted, angry, and neutral) has a plausible explanation for its context and a corresponding response. }
    \label{fig:emovdb}
\end{figure*}

To answer this question, we introduce \ourbench, a benchmark designed to assess the \textbf{empathic responsiveness and judgment} capabilities of OLMs. It comprises 1,885 English instances, each pairing an audio utterance with a concise textual context. It overcomes the limitations of existing empathetic dialogue evaluations, which lack multimodality (see Fig.~\ref {fig:input-modes}) and parallel comparisons (over varying contexts and acoustic tones). There exhibits two novel design strategies: (\textbf{i}) \textit{ Contextual Variation}, where distinct background contexts are applied to the same utterance to alter its pragmatic meaning (Fig.~\ref {fig:gigaspeech}); and (\textbf{ii}) \textit{Tonal Variation}, where identical contexts are paired with the same utterance delivered in varying emotional tones to shift user intent (Fig.~\ref{fig:emovdb}). To \emph{the best of our knowledge, \ourbench is the first benchmark to jointly examine linguistic and paralinguistic empathy via parallel contexts and acoustic tonal variations}.



We then conducted an extensive evaluation on \ourbench, benchmarking current state-of-the-art OLMs on their ability to \textit{generate empathetic responses} and \textit{judge generations from other models} compared to human evaluators. 
This assessment covered seven dimensions designed to comprehensively measure both linguistic and paralinguistic features: \textit{modality reliance}, \textit{naturalness}, \textit{coherence}, \textit{supportiveness}, \textit{discrimination}, and \textit{delivery}. 

Experimental results demonstrate that OLMs trained with audio output capabilities generally outperform those limited to text-only output. GPT produced the most human-like and empathetic narratives, followed closely by Qwen-Omni. These findings underscore the value of integrating text and audio for learning empathy. However, while OLMs align with human judgment on coarse-grained tasks, they remain unreliable for fine-grained emotional evaluation. Specifically, the models lack paralinguistic empathy, often delivering flat speech, and OLM-based judges diverge significantly from human perception regarding emotional prosody. Consequently, generating and evaluating expressive empathetic tone remains an open challenge.




\section{Related Work}
\label{sec:related}


\paragraph{Empathy and Emotional Intelligence. }
Empathy is widely defined as the ability to perceive, understand, and respond to another individual's emotional states.
It is fundamental for satisfactory and effective conversational communication~\cite{szalita1976empathy}. 
Empathy in dialogue has been studied in text through datasets and frameworks that annotate empathic responses and strategies. \textsc{EmpatheticDialogues} introduced 25k conversations grounded in emotional situations for empathic response generation~\cite{rashkin2019empathetic}. Work in mental-health support provides theory-grounded taxonomies and labelled corpora to detect and analyse empathy mechanisms in text~\cite{sharma2020computational}. With the development of LLMs, automatic evaluation of response empathy is available~\cite{zhang2025sentient}. 
For generated audio responses, it remains confined to the text modality empathy evaluation, converting the audio content into text~\cite{wang2025emotion}. 
In contrast, we evaluate the potential of OLMs as automatic judges directly on the audio response, in order to study empathy from both linguistic and paralinguistic perspectives.

\begin{table*}[t]
    \centering
    \begin{tabular}{l|ccccr}
    \hline
       \textbf{Subset}  &  \textbf{Source}  & \textbf{Language} & \textbf{Core Contribution} & \textbf{Length} \\
    \hline
       \meld  & TV Series  & Colloquial  &  Daily Conversational Emotion & 18, 30 \\
       \emovdb & Studio recording  & Literary & Explicit Emotional Range & 10, 20 \\
       \giga &  Streaming content  & Formal  &  Domain Diversity, Acoustic Variability & 22, 17 \\
    \hline
    \end{tabular}
    \vspace{-0.5em}
    \caption{Contribution of each data source to \ourbench. \emovdb and \meld primarily contain audios with clear emotions. Length is the average words of the utterances and generated contexts in \ourbench. }
    \vspace{-0.5em}
    \label{tab:dataset-features}
\end{table*}

\paragraph{Paralinguistic Research and Audio Cues.}
Human language is generally split into the verbal channel (the semantic content or chosen words) and the vocal channel (the non-semantic delivery~\cite{trager1958paralanguage}. Paralinguistic features belong to this vocal channel, conveying emotional meaning through prosody, which is variations in pitch, loudness, timbre, speech rate, and pauses.
These paralinguistic cues are indispensable for empathetic communication~\cite{loveys2021effects} because they can drastically alter the meaning of verbal content, such as in the case of sarcasm, or reveal emotional states that text transcription alone cannot capture. 
MFCCs (Mel-Frequency Cepstral Coefficients)~\cite{mfcc}, for instance, capture the characteristics of the sound that are most discernible to the human ear.
\meld~\cite{poria-etal-2019-meld} and IEMOCAP~\cite{busso2008iemocap} are widely used benchmark for SER. 
\emovdb~\cite{adigwe2018emotional} is an emotional speech dataset used for synthesis and generation purposes, offering varied content in different tones delivered by both male and female voices.
Furthermore, specialised datasets, such as VocalSound~\cite{gong2022vocalsound}, provide collections of non-lexical human vocalisations, including laughter and sighs, to train and evaluate models across a broader range of human-voice acoustics.
For multimodal dialogue, the evaluation on generated audio quality is conducted by human listeners rating the naturalness or quality of the speech~\cite{zhang2025towards, xu2025qwen2, du2024cosyvoice}, using Mean
Opinion Score (MOS)~\cite{viswanathan2005measuring}.
However, OLMs' sensitivity to paralinguistic perception, such as tone and emotions, remains underexplored. 
\ourbench fills the gaps by setting the dialogue with both text and human audio, thereby evaluating the multi-modal empathetic conversation ability and investigating the feasibility of automatic speech evaluation using OLMs.

\section{\ourbench Benchmark}

\subsection{Benchmark Design}

Empathy broadly refers to our reaction to the experiences of others. It has two major components: 1)~\textbf{Cognitive empathy},  defined as the process of understanding another person’s perspective\footnote{In real human interaction, most daily conversations are informational or neutral. In these settings, empathy is expressed not through emotional mirroring but through helpfulness, perspective-taking, alignment with the interlocutor’s goals, and context-appropriate support. }
and 
2)~\textbf{Affective empathy}, also called emotional empathy, defined as an observer’s emotional response to the affective state of others~\cite{davis1983measuring}.

\ourbench for evaluating OLMs' empathy is constructed from three complementary datasets: \meld~\cite{poria-etal-2019-meld}, \giga~\cite{chen2021gigaspeech}, and \emovdb~\cite{adigwe2018emotional}. 
They span the critical dimensions of real-world communication: contextual dialogue, domain diversity, and explicit emotional expression.\footnote{We exclusively employ human speech to ensure natural paralinguistic variability (e.g., accents, prosody), a complexity absent in stable synthesised voices.}
The contexts are constructed in reverse according to the Utterance. 
The features of each data source are present in Table~\ref{tab:dataset-features} and the composition is listed in Table~\ref{tab:bench-stat}.
It is classified into two types (985 chatting + 900 asking for help) and 6 topics (Appx. Table~\ref{tab:bench-type-topic}).

\begin{table}[t]
    \centering
    \begin{tabular}{l|rrr}
    \hline
        \textbf{Source} & \textbf{\#Audio} & \textbf{\#Context} & \textbf{Total} \\
    \hline
         \meld & 281 & ($\times$3) 843 & 843 \\
         \giga & 303 & ($\times$2) 606 & 606 \\
         \emovdb & ($\times$4) 436 & 109 & 436\\
    \hline
        \textbf{Total} & 1,020 & 1,558 & 1,885 \\
    \hline
    \end{tabular}
    \vspace{-0.5em}
    \caption{Statistics of \ourbench. There are 1,885 English instances, each pairing an audio utterance with a textual context. ($\times n$) means the number at the cell is $n$ times to the neighbouring number at the same row.   
    }
    \vspace{-0.5em}
    \label{tab:bench-stat}
\end{table}

\ourbench covers two complementary evaluation tasks, as illustrated in Fig.~\ref{fig:gigaspeech} and~\ref{fig:emovdb}:

\paragraph{1) Same utterance, different contexts.} 
The first type of data focuses on the same utterance audio used in different contexts (\textbf{A–Cs}), where background framing shifts its interpretation. For example, the statement ``\textit{Then just give him some money}'' could occur in a discussion about supporting a struggling relative, implying personal aid, or in a conversation about backing a political candidate, suggesting financial sponsorship. Despite the identical wording, the empathic strategies required are very different. (See the example in Fig.~\ref{fig:cases-giga}).

\paragraph{2) Same context, different utterances.} 
The second type of data involves the same context and utterance \textit{audio} in different emotional tones (\textbf{C-As}). Although the literal words remain identical, the affective delivery changes the underlying meaning. For instance, an angry ``\textit{Sometimes you really need to quarrel to solve problems}'' conveys frustration and a need for validation, whereas the same sentence spoken calmly suggests a pragmatic and constructive view of conflict. Models should recognise the emotional signal and adjust their response style accordingly. (See the example in Fig.~\ref{fig:cases-emovdb}).

\subsection{Benchmark Construction}
Our benchmark is constructed in three steps:


\paragraph{1) Audio Filtering.} Because our benchmark focused on empathy evaluation, we filtered out scenarios of (i) Physical actions, (ii) Mathematical reasoning or complex task completion, and (iii) Targeting real people. More details are in Appx.~\ref{sec:appendix:construction}.

\paragraph{2) Contexts Generation.}   
We adopted GPT to frame different contexts for the retained data. 

$\bullet$ \textbf{A–Cs} evaluate the ability to associate a specific context. \meld and \giga both have one specific audio for each utterance. \meld are daily short colloquial conversations that could fit in various contexts. \giga are more domain-specific and formal ones, restricting their flexibility. Therefore, we set 3 contexts for each \meld audio and 2 contexts for each \giga audio. 
    
$\bullet$ \textbf{C–As} evaluate the hearing ability of models. Different tones reflect the speaker's different attitude towards the same context, which requires models to infer the speaker's role or standpoint given a specific context. \emovdb recorded utterances with different explicit emotions. Here, we set only one context for all the different tones.


\paragraph{3) Quality Validation.} 
To further validate the data quality of \ourbench, we first prompted GPT to generate a plausible explanation for why the context and utterance are coherent and provide a reference response. 
Then, human annotators manually verified that each constructed context is coherent.\footnote{More annotation details are discussed in Appx.~\ref{sec:appendix:construction}.}
Specifically, they checked:
(i) The context plausibly leads to the audio. 
(ii) For A–Cs, the contexts are semantically different; For C-As, all the emotions are explainable in such contexts.



\subsection{Evaluation Metrics}
We evaluate from two aspects: \textbf{linguistic} and \textbf{paralinguistic}.
\textit{modality reliance}, \textit{naturalness}, \textit{coherence}, \textit{supportiveness}, and \textit{discrimination} are evaluated as linguistic features.
For paralinguistic, we focus on \textit{delivery}.\footnote {We also attempted to evaluate paralinguistic \textit{emotions}, yet it cannot work. We discussed details in the Limitation section.}  
Model judges are shown in Table~\ref{tab:placeholder} and more details in Appx.~\ref{sec:appendix:metric} and~\ref{sec:appendix:scores}.


\begin{table}[h]
    \centering
    \begin{tabular}{c|c|c}
    \hline
        \textbf{Judge} & \textbf{Modality} & \textbf{Metrics} \\
    \hline
        GPT-5 & Text &  M.R., Nat., Disc.\\
        OLMs & Audio & Coh., Sup., Delivery \\
    \hline
    \end{tabular}
    \vspace{-0.5em}
    \caption{The automatic judges for evaluation metrics.}
     \vspace{-0.5em}
    \label{tab:placeholder}
\end{table}

$\bullet$ \textbf{Modality Reliance} (M.R.): \textit{Which modality does the model rely on to generate the response? }

$\bullet$ \textbf{Coherence} (Coh.): \textit{Is the response logically consistent to the context and utterance?} (semantic check for 
(Good, Fair, Poor))
~\cite{ickes2000difficulty}.

$\bullet$ \textbf{Naturalness} (Nat.): \textit{How natural is the response?} More human-like responses have higher scores with the range (1-4)~\cite{kuhne2020human}.

    
$\bullet$ \textbf{Discrimination}  (Disc.): \textit{Across different contexts/tones, do the model responses vary?} Higher scores are for those tailored for context/tone with the range (NA, 1–6)~\cite{ickes2000difficulty}.
            
$\bullet$ \textbf{Supportiveness} (Sup.) : \textit{Imagine yourself in this situation. How supportive does the response feel? } Here, we consider acknowledgment of feelings, perspective-taking, supportive intent (comfort/encouragement/help), and non-judgmental language for (Good, Fair, Poor)~\cite{rogers1957necessary}.
    

$\bullet$ \textbf{Delivery}: \textit{How supportive the response sounds?} Here we consider the tone of voice, pacing, pauses, etc., for (Good, Fair, Poor)~\cite{burleson2009revisiting, loveys2021effects}.  

\section{Experimental Settings}

\paragraph{Baselines.} 
1) Qwen3-Omni~\cite{xu2025qwen3omnitechnicalreport},
2) Qwen2.5-Omni~\cite{xu2025qwen2},
3) Qwen2-Audio~\cite{chu2024qwen2},
4) Qwen-audio~\cite{chu2023qwen},
5) SALMONN~\cite{tangsalmonn},
6) LLaMA~\cite{fang2025llama},
7) Flamingo~\cite{ghosh2025audio},
8) Baichuan~\cite{li2025baichuan}, and 
9) GPT~\cite{hurst2024gpt}. 
(See Appx.~\ref{sec:appendix:baselines}).

\paragraph{Human Annotation. }
For empathy judgement, we compared OLMs' results to human's. 
Audio responses were sampled evenly from each OLM. 
For each model pair, their responses to the same instance were evaluated by the same annotator to ensure consistent judgment; also, inter-annotator consistency was computed based on models' ranking within each pair. Details are described in Appx.~\ref{sec:appendix:annotation}.


\paragraph{Consistency Measurement. }
Because empathetic scores are relatively subjective, for the \textit{Inter-annotator consistency}, annotations are considered consistent if any of the following conditions hold:
(i) the paired models receive different scores but the same rank order from both annotators;
(ii) the paired models are tied in score by both annotators; or
(iii) one annotator records a tie, and the other reports a score difference $\leq$ 1.  
For \textit{Human–OLM judge consistency}, each model response is first assigned the average score of its two human annotators. The resulting average rank for each model pair is then compared against the OLM-judge ranking, following the aforementioned conditions.

\begin{table}[!h]
  \centering
  \setlength{\tabcolsep}{5pt}
  \begin{tabular}{l|rccrc}
    \hline
     
     \textbf{Model} & T & A & T+A & F & Norm \\
    
    \hline
    
    GPT & 0 & 18 & 82 & 0 
     & \textbf{0.91} \\
    
    LLaMA  & 1 & 36 & 62 & 1 
     & 0.81 \\

    Baichuan  & 6 & 34 & 59 & 1 
     & 0.79 \\


    Qwen3-Omni & 1 & 24 & 74 & 0 
     & 0.87 \\
    Qwen2.5-Omni & 0 & 25 & 74 & 0 
     & 0.87 \\

    \hline
    Qwen2-Audio & 4 & 18 & 52 & 25
     & 0.63 \\
    Qwen-Audio & 5 & 18 & 65 & 12 
     & 0.77 \\

    SALMONN & 13 & 27 & 57 & 3 
     & 0.77 \\

    Flamingo & 10 & 36 & 47 & 7 
     & 0.70\\

    \hline
  \end{tabular}
\vspace{-0.5em}
  \caption{Evaluation on input modality reliance. The values indicate the proportion (\%) of responses based on text only (T), audio only (A), both text and audio (T+A), or failure to respond (F). Normalised score (Norm) is the sum of the value T+A and the average of T and A.}
  \vspace{-1em}
  \label{tab:input-modality}
\end{table}

\section{Results on Omni-Models as Responders}

\begin{table*}[!th]
  \centering 
  \setlength{\tabcolsep}{2pt}
  \begin{tabular}{cl|cccccc|cc|cc}
    \hline
    Audio & \multirow{2}{*}{\textbf{Model}} & \multicolumn{2}{c}{\meld}  & \multicolumn{2}{c}{\giga} & \multicolumn{2}{c}{\emovdb} & \multicolumn{2}{|c}{Overall} & \multicolumn{2}{|c}{Normalised Scores}  \\
    Output  &  & \textcolor{white}{..}Nat. & Disc. & \textcolor{white}{..}Nat.\textcolor{white}{..} & Disc. & Nat. & Disc. & Nat. & Disc. & Nat. (Rank) & Disc. (Rank) \\
    \hline
    
    \Checkmark & GPT & \textbf{3.95} & 4.85 & \textbf{3.93} & 4.38  & \textbf{3.97} & 2.17 &  \textbf{3.95} & 4.22 
    & \textbf{0.98} (1) & 0.64 (6) \\ 
    
    \Checkmark & LLaMA  & 3.88 & 4.81  & 3.77 & 4.79 & 3.88 &  2.70 & 3.85 & 4.47 
    & 0.95 (4) & 0.69 (4) \\

    \Checkmark & Baichuan  & 3.90 &  \textbf{5.15} & 3.81 &  4.95  & 3.94 & 3.08 & 3.88 & 4.74
    & 0.96 (2)  & 0.75 (2) \\
    
    \Checkmark & Qwen3-Omni & 3.61 & 5.14 & 3.88 & \textbf{5.02} & 3.93 & 2.77 & 3.77 & 4.71 
    & 0.92 (5) & 0.74 (3) \\
    
    \Checkmark & Qwen2.5-Omni & 3.87 &  4.38 & 3.87 & 4.63 & 3.94 & 2.34 & 3.89 & 4.17 
    & 0.96 (2) & 0.63 (7) \\

\hline
    - & Qwen2-Audio & 3.01 & 4.63 & 2.32 & 3.85 & 3.18 & 2.93  & 2.82 & 4.00 
    & 0.61 (7) & 0.60 (8) \\
    
    - & Qwen-Audio & 2.25 & 4.78 & 2.46 & 5.01  & 2.48 & \textbf{4.63}  & 2.37 &  \textbf{4.86} 
    & 0.46 (8)  & \textbf{0.77} (1) \\

    - & SALMONN & 3.25 & 4.61 & 3.18 & 4.66 & 3.31 & 3.09 & 3.24 & 4.39 
    & 0.75 (6) & 0.68 (5) \\

    - & Flamingo & 2.40 & 4.10  & 2.31 & 3.79  & 1.95 &  2.66 & 2.27 & 3.73 
    & 0.42 (9) & 0.55 (9) \\

    \hline
  \end{tabular}
  \vspace{-0.5em}
  \caption{Naturalness (Nat.) measures how human-like the responses are (ranges 1-4 and the higher the more human-like). Discrimination (Disc.) assesses the model’s ability to distinguish responses across varying contexts (\meld \& \giga) or audio tone (\emovdb) (ranges 1-6, and the higher the better). They both are measured on text and evaluated by GPT-5-mini (with manual checking) following the text evaluation practice \cite{fu-etal-2024-gptscore}.}
  \vspace{-0.5em}
  \label{tab:linguistic}
\end{table*}

We first discuss model responses with empathy.
Table~\ref{tab:input-modality} shows \textit{Modality Reliance} of baselines,  Table~\ref{tab:linguistic} their \textit{Naturalness and Discrimination}, and 
Table~\ref{tab:paralinguistic}  \textit{Delivery} evaluated by human and OLM-judges. We observe that:
(1) OLMs with audio outputs outperform text-only-output models. GPT and Qwen’s Omni series represent the top-tier audio synthesis quality. These indicate the benefits of multimodal integration in learning empathy.
(2) Models' performance varies on Discrimination and Naturalness, indicating they may be decoupled abilities. 

$\bullet$ \textbf{Modality Reliance.}
As shown in Table~\ref{tab:input-modality}, all OLMs primarily rely on both audio and text modalities (over 50\%).
GPT demonstrates the best multimodal integration (normalised to 0.91), followed by Qwen's omni series (0.87).
Earlier audio analysis models exhibit higher failure rates and weaker multimodal grounding, i.e., Qwen's audio series.

$\bullet$ \textbf{Naturalness.}
Among OLMs, GPT achieves the highest overall Naturalness (normalised to 0.98), closely followed by Baichuan and Qwen2.5-Omni (both 0.96). It suggests that recent OLMs can produce responses better resembling human conversational style.
In contrast, Flamingo and the Qwen-Audio series exhibit substantially lower Naturalness, often providing analytical or descriptive content rather than natural, conversational replies.

$\bullet$ \textbf{Discrimination.}
In contrast, Qwen-Audio attains the highest Discrimination score (normalised to 0.77), particularly on the \emovdb subset. It indicates strong sensitivity to varying emotional tones of input audio. However, this sensitivity is primarily due to analytical variation (not communicating), as evidenced by its poor Naturalness.

$\bullet$ \textbf{Delivery.}
Only OLMs with audio outputs are evaluated for the paralinguistic metric. Both Human and OLM-judges consistently rank GPT, Qwen3-Omni, and Qwen2.5-Omni as the top-tier models in paralinguistic quality, indicating superior control over prosody and vocal delivery compared to LLaMa and Baichuan.


\begin{table*}
  \centering
  \begin{tabular}{l|ccc|ccc|ccc|ccc}
    \hline
    \multirow{2}{*}{\textbf{Judge}} & \multicolumn{3}{c|}{\meld (\%)}  & \multicolumn{3}{c|}{\giga (\%) } & \multicolumn{3}{c|}{\emovdb (\%) } & \multicolumn{3}{c}{Average (\%) } \\
    &  Coh. & Sup. & Para. & Coh. & Sup. & Para. & Coh. & Sup. & Para. & Coh. & Sup. & Para.  \\
    \hline
    Human & 89.7 & 89.7 & 87.9
    & \textbf{95.0} & 90.0 & 80.0
    & \textbf{82.5} & \textbf{88.8} & \textbf{88.8}
    &  \textbf{87.6} & \textbf{89.3} & 86.5 \\
    \hline
    GPT & \textbf{94.1} & 85.3 & \textbf{88.2} & 87.5 & 83.3 & \textbf{100.0} & 77.1 & 75.0 & 83.3 & 84.9 & 80.2 & \textbf{88.7} \\

    Q-Omni & 87.9 & \textbf{91.4} & 86.2 & 92.5 & \textbf{95.0} & 82.5 & 57.5 & 83.8 & 75.0 & 75.3 & 88.8 & 80.3 \\ 

    Q-Audio  & 58.6 & 60.3 & 75.9 & 57.5 & 65.0 & 72.5 & 57.5 & 51.3 & 56.3 & 57.9 & 57.3 & 66.3  \\

    \hline
  \end{tabular}

  \caption{The consistency between human annotators and model judges on coherence (Coh.), supportiveness (Sup.), and Delivery (Para.), respectively. Model judges are GPT-4o-audio-preview (GPT), Qwen2.5-omni (Q-omni), and Qwen2-Audio (Q-Audio). The row of Human is the annotation consistency between two human annotators.}
  \label{tab:judge-alignment}
\end{table*}

\begin{table*}[!t]
\centering
  \setlength{\tabcolsep}{2.5pt}
\begin{tabular}{l|ccccc|ccccc|ccccc}
\hline
\multirow{2}{*}{\textbf{Model}} & \multicolumn{5}{c|}{\meld}         & \multicolumn{5}{c|}{\giga}   & \multicolumn{5}{c}{\emovdb}       \\
                       & GPT  & Q3   & Q-O  & Q-A  & H    & GPT  & Q3   & Q-O  & Q-A  & H    & GPT  & Q3   & Q-O  & Q-A  & H    \\
\hline

GPT                    & \textbf{0.65} & \textbf{0.99} & \textbf{0.96} & 0.14 & 0.77 & \textbf{0.56} & \textbf{0.98} & \textbf{0.93} & 0.06 & 0.81 & \textbf{0.59} & \textbf{1.00} & 0.96 & 0.09 & 0.55 \\
Qwen3-omni             & 0.56 & 0.84 & 0.81 & 0.95 & \textbf{0.94} & 0.53 & 0.95 & \textbf{0.93} & 0.97 & \textbf{0.85} & 0.56 & 0.96 & \textbf{0.96} & 0.94 & \textbf{0.86} \\
Qwen2.5-Omni           & 0.61 & 0.95 & 0.92 & 0.14 & 0.93 & 0.55 & 0.97 & 0.88 & 0.09 & \textbf{0.85} & 0.57 & 0.97 & 0.95 & 0.10 & 0.80 \\
\hline
LLaMA                  & 0.54 & 0.84 & 0.78 & 0.13 & 0.67 & 0.51 & 0.84 & 0.75 & 0.09 & 0.52 & 0.53 & 0.86 & 0.82 & 0.06 & 0.34 \\
Baichuan               & 0.49 & 0.74 & 0.72 & 0.15 & 0.64 & 0.51 & 0.73 & 0.67 & 0.08 & 0.61 & 0.51 & 0.76 & 0.79 & 0.08 & 0.53\\
\hline
\end{tabular}
\caption{Evaluation on paralinguistic features: \textit{Delivery}. For easy reading, we present normalised scores from a 3-point Likert scale (the higher, the better). The columns show the judges evaluating models in rows. Judges include GPT-4o-audio-preview (GPT), Qwen3-Omni (Q3-O), Qwen2.5-Omni (Q-O), Qwen2-Audio (Q-A), and Human (H). The boldface numbers indicate the best ones according to the original scores, listed in Table~\ref{tab:paralinguistic-appendix}, Appx.~\ref{sec:appendix:good-fair-poor}.}
  
  \label{tab:paralinguistic}
  
\end{table*}

\section{Results on Omni-Models as Evaluators}

We then discuss how OLMs evaluate empathy from generated audio.
Tables \ref{tab:paralinguistic} and \ref{tab:judge-alignment} show the results.\footnote{The Coherence and Supportiveness scores are in Appx.~\ref{sec:appendix:good-fair-poor}.}

\paragraph{Alignment with Human Judgment. }
In Table~\ref{tab:judge-alignment}, GPT exhibits the highest overall consistency with human annotators across coherence (84.9\%) and delivery (88.7\%).\footnote{For Delivery, level-samples are provided in prompts, which suggests GPT has stronger in-context-learning ability.} It closely approaches human-human agreements.
Qwen2.5-Omni shows competitive performance in terms of supportiveness (average 88.8\%, comparable to human-human interactions).
These imply that the best OLMs have the potential for coarse-grained automatic empathy evaluation from audio.
It again highlights the benefits of more generic, multimodal capabilities in understanding empathy.
In contrast, Qwen2-Audio yields substantially lower agreement across all dimensions, confirming that earlier audio-analysis–oriented models are not effective here.


\paragraph{OLMs' Evaluation Consistency}
Table~\ref{tab:paralinguistic} shows the ratings from OLM- and human-judges over delivery. Qwen2-Audio (Q-A) exhibits polarised scoring and a marked bias toward Qwen3-O, hindering reliable rank differentiation for scores exceeding 2. In contrast, other OLM-judges consistently distinguish Qwen-Omni models and GPT from lower-tier models, aligning well with human judgment. This demonstrates the potential of OLM-judges to effectively evaluate coarse-grained delivery. However, regarding fine-grained delivery, a divergence emerges: while human judges prefer the Qwen-Omni series over GPT, OLM-judges favour the opposite. These findings suggest that the ability of OLMs to discern fine-grained superiority in delivery is not yet aligned with human preference.




\begin{figure}[]
    \centering
    \includegraphics[width=\linewidth]{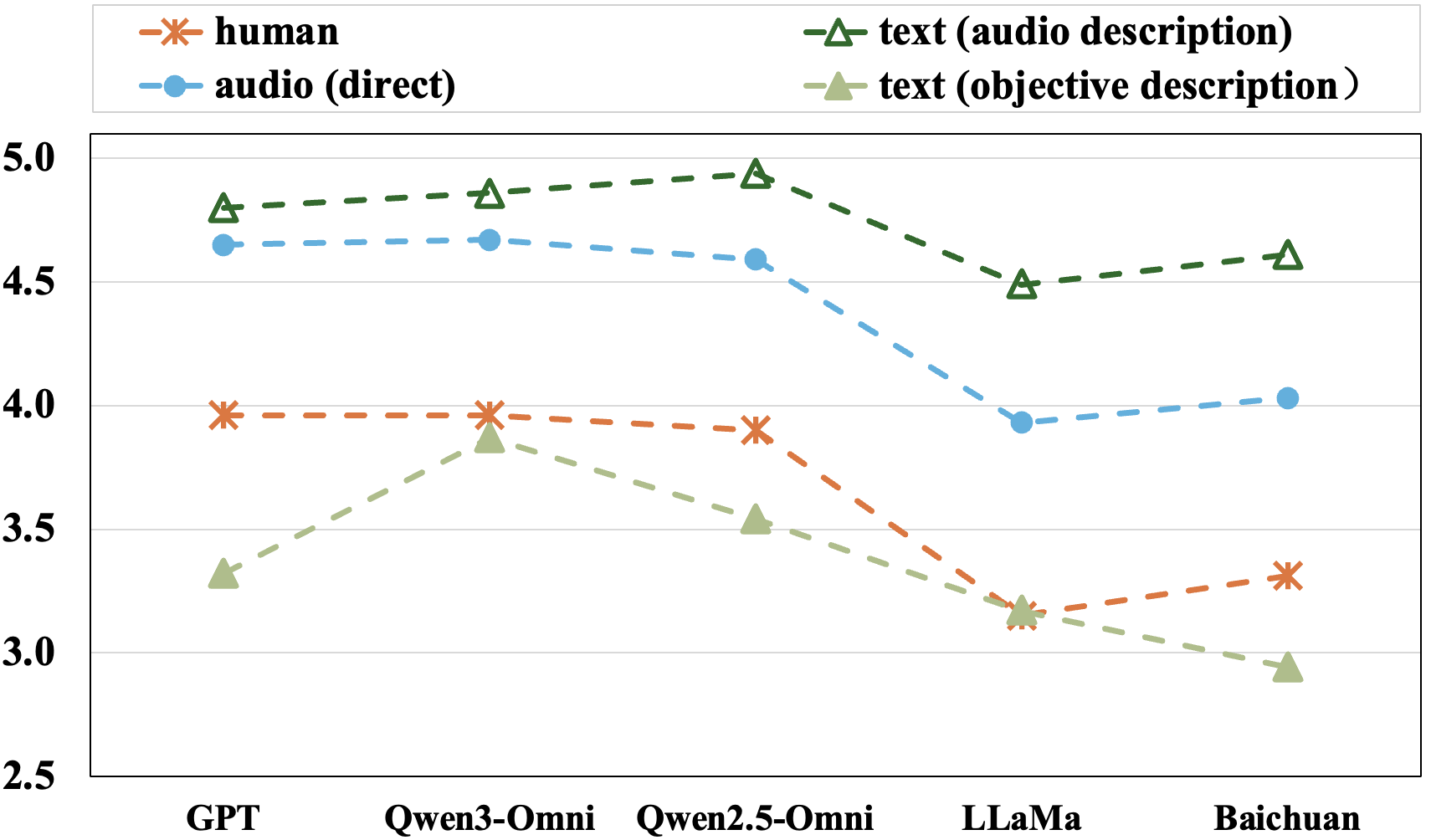}
     \vspace{-1em}
    \caption{Average 5-point paralinguistic (delivery) evaluation (y-axis) for model on x-axis. Higher is better. For judges, stars indicate human evaluation, while circles indicate OLM-judges using direct audio. Empty and filled triangles denote OLM-judges using additional contexts of the original and objective audio caption, respectively. }
    \vspace{-1em}
    \label{fig:fine-grained}
\end{figure}

\begin{figure*}[t]
    \includegraphics[width=\linewidth]{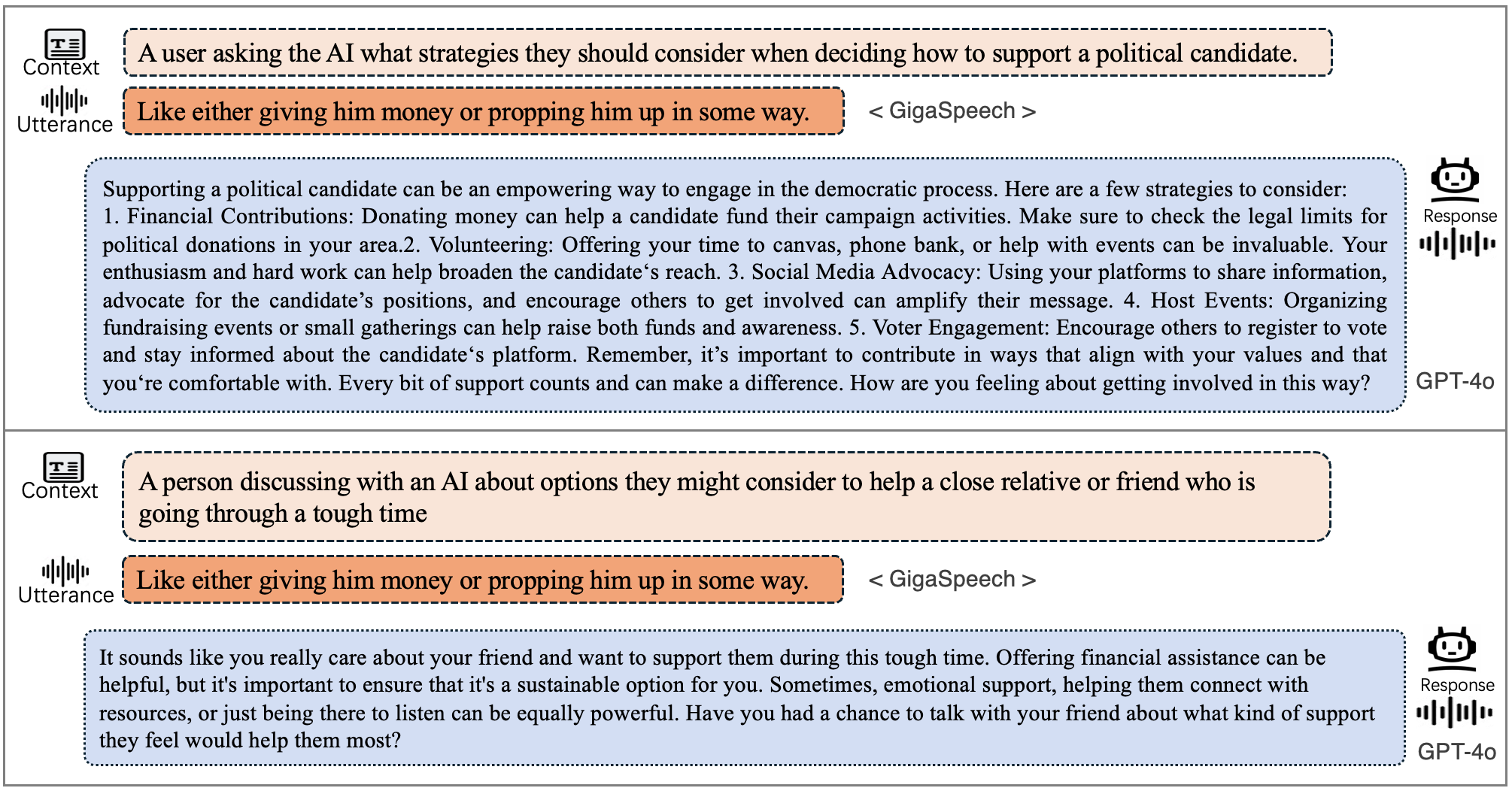}
    \vspace{-1.5em}
    \caption{The example of context-variant responses generated by \textit{GPT-4o}. Its responses are typically long and adhere to a standardised structural format: first, a situation analysis, then suggestions, and finally a question.}
     \vspace{-0.5em}
    \label{fig:cases-giga}
\end{figure*}
\begin{figure*}[t]
    \includegraphics[width=\linewidth]{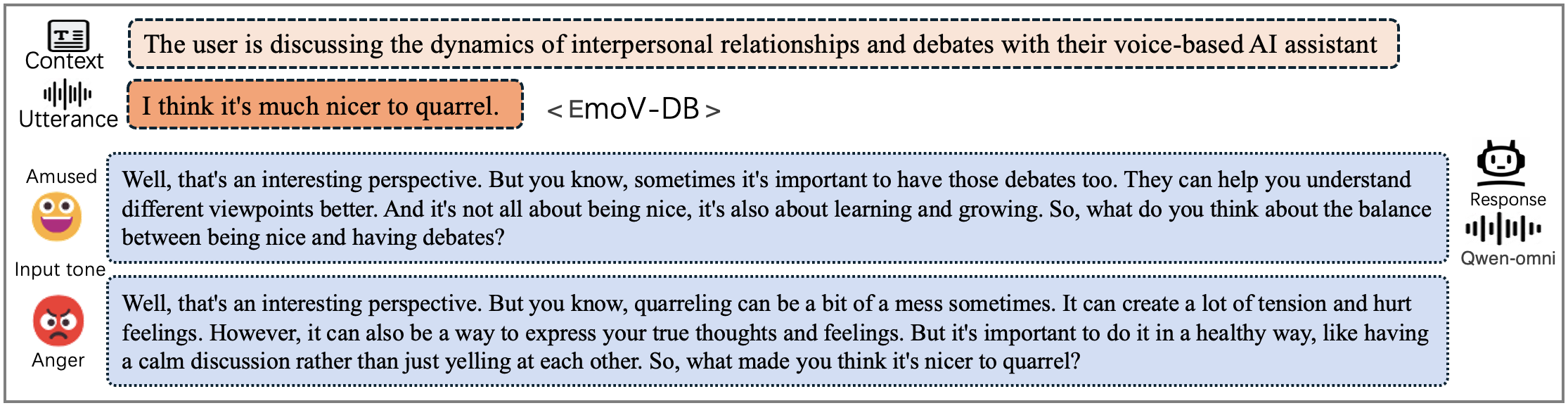}
     \vspace{-1em}
    \caption{ The example of tone-variant responses generated by \textit{Qwen2.5-Omni}. The model shows a distinct stylistic pattern, frequently starting with ``\textit{well}'' and adopting a reflective tone marked by phrases such as ``\textit{but you know}''.}
     \vspace{-0.5em}
    \label{fig:cases-emovdb}
\end{figure*}

\section{Further Discussion}

\paragraph{Quantitative  Analyses of Paralinguistics.}
We have highlighted the challenges in fine-grained evaluation of paralinguistic delivery. To provide further insight, we investigate whether textual descriptions of audio (captions) can aid in assessing paralinguistic empathy. We utilised Qwen3-Omni for audio captioning and conducted a fine-grained evaluation, expanding the scoring metric from a 3-point to a 5-point scale to capture greater nuance.
Results are shown in Fig.~\ref{fig:fine-grained} with details in Appx.~\ref{sec:appendix:fine-grained}.

Direct audio judging (circles) aligns with the general ranking order of human (stars) but achieves only 43\% consistency with human judgements. Incorporating the original audio captions (empty triangles) exacerbates the issue, as the captions tend to be biased and overly positive regarding machine delivery (e.g., ``\textit{The speaker's voice is calm, gentle, and empathetic...}''). Furthermore, although prompting the OLM to emphasise objective descriptions (filled triangles) brings the results closer to human evaluation, it tends to introduce a negative bias. It implies that captions can partially help, yet their effectiveness is highly sensitive to their quality. 
Moreover, these results demonstrate that textual audio captions, whether prompted for objectivity or not, are not faithful proxies for auditory perception. It is because nuanced tones and implicit cues can be easily overlooked and excluded in captions. As a result, captions cannot substitute for direct audio evaluation, particularly when assessing emotional delivery. It highlights a critical limitation of current OLM evaluators and underscores the necessity for a more dedicated audio paralinguistic modelling.


\paragraph{Qualitative Analyses.} We further interviewed the annotators and found GPT-4 to be perceived as the most human-like system, followed by Qwen-Omni. While providing empathetic content, Qwen-Omni tends to append template-style sentences, resulting in less discrimination. 
Baichuan shows more stable conversational behaviour than LLaMA, whereas LLaMA occasionally generates sarcastic responses rather than uniformly supportive ones.

Finally, we present a qualitative analysis of the generated samples. While generally lacking strong expressive prosody, interestingly, the models display distinct stylistic characteristics. GPT-4o offers fast, structured responses with smooth, standardised delivery (see Fig.~\ref{fig:cases-giga}). Qwen2.5-Omni adopts a reflective tone, often initiating with ``\textit{well}'' at a moderate pace (see Fig.~\ref{fig:cases-emovdb}). While Qwen3-Omni occasionally generates vivid, emotional audio, it suffers from instability. Baichuan shows lower consistency, characterised by evident fluctuations in accent. These findings highlight a persistent gap in achieving truly human-like, empathetic generation.




\section{Conclusion}
We have presented \ourbench to evaluate empathetic response generation and judgments in OLMs. By employing novel context- and tone-variant settings, we found that although top-tier OLMs exhibit strong linguistic empathy, they still face significant challenges in mastering fine-grained paralinguistic delivery. Moreover, our results show that textual audio captions cannot replace direct audio analysis, underscoring the urgent need for audio-native evaluators to develop genuinely empathetic AI systems.

\clearpage
\section*{Limitations}
Our work provides an initial exploration into the EQ of OLMs, but it is subject to several limitations that suggest future research directions:

Language and Cultural Scope: \ourbench is constructed exclusively in English. Empathy is highly language- and culture-dependent, meaning the emotional cues, appropriate supportiveness strategies, and even the naturalness of delivery can vary drastically across different linguistic groups. 

Coarse-Grained Metric: Our metrics use a coarse-grained categorical rating ("Good," "Fair," "Poor"). 
While evaluating audio quality using OLM judges has high consistency with coarse-grained metrics, the complexity of assessing the nuanced emotional intent and consistency of the model-generated audio output remains challenging.

Emotion evaluation: The lack of expressive variation in the synthesized audio results in mostly neutral emotion in responses. The GPT/Qwen models primarily rely on textual analysis to judge emotion, rating various emotion distributions. So far, the acoustic quality limitation of the synthesized speech prevents the evaluation results from reliably showing a disparity in emotional rendering between the tested systems.

\section*{Ethical Considerations}
The dual use of highly empathetic OLM capabilities presents the primary risk: the generated supportive responses could be exploited for manipulation, phishing, or disinformation. Furthermore, the study raises concerns about fairness and bias due to the potential for emotional overgeneralization or bias confirmation stemming from the underlying data. Finally, processing a user's sensitive emotional state from audio highlights significant privacy risks. Our work is confined to research, with future deployment requiring strong safeguarding mechanisms and adherence to privacy-by-design principles.

Additionally, to ensure ethical annotation data collection, all human annotators (student helpers) involved in model response evaluation were compensated fairly. Annotators were paid at a standard university rate of 16 USD per hour.

\bibliography{acl}

\appendix

\section{Details of Utilized Benchmark}
\label{app:benchmark details}

The comparisons are listed in Table~\ref{tab:benchmarks} and~\ref{tab:dataset-features-apx}.

1) \meld (Multimodal EmotionLines Dataset), sourced from TV series Friends and was labeled for Emotion Recognition in Conversation. The dialogues are daily communication among family, friends, and neighbourhood rather than academic or knowledge. 

2) \giga, compiled from diverse YouTube and Podcast content,  tests the system's ability to handle relatively formal and terminology across topics, thereby broadening the EQ assessment beyond simple daily conversation.

3) \emovdb (Emotional Voices Database) provides the acoustically controlled emotional ground truth. \emovdb’s studio-recorded speech offers clean, high-fidelity examples of explicit core emotions. This inclusion is crucial for isolating pure paralinguistic cues and ensuring the system can accurately discriminate core acoustic-emotional features, independent of the complexities introduced by dialogue context. 

The topic distribution are listed in Table~\ref{tab:bench-type-topic}.

\section{Details of Models}
\label{sec:appendix:baselines}
Tested baseline models are listed in Table~\ref{tab:baseline-comparison}. 

1) \textbf{Qwen3-Omni-30B-A3B-Instruct}~\cite{xu2025qwen3omnitechnicalreport} and 
2) \textbf{Qwen2.5-Omni-7B}~\cite{xu2025qwen2} have an end-to-end architecture that processes text, images, audio, and video inputs and generates simultaneous text and speech outputs. It is designed for streaming, low-latency, and deployment on edge devices. 

3) \textbf{Qwen2-Audio-7B-Instruct}~\cite{chu2024qwen2} and 4) \textbf{Qwen-audio-chat}~\cite{chu2023qwen} are similar series, both designed for audio chat (with text output only) and audio analysis, such as music analysis or sound recognition.

5) \textbf{SALMONN-7B}~\cite{tangsalmonn} is capable of emotion recognition, speaker verification, and music and audio captioning beyond traditional tasks.

6) \textbf{LLaMA-Omni2-7B}~\cite{fang2025llama} focuses on creating a real-time spoken chatbot by streaming speech synthesis. Although it can only process audio and text modalities, the "Omni" suggests broader multimodal capabilities in the LLaMA series.

7) \textbf{audio-flamingo-3}~\cite{ghosh2025audio} advances understanding and chain-of-thought-type reasoning across speech, along with multi-turn, multi-audio chat ability. Code for Flamingo Streaming-TTS pipeline has not been released yet. 

8) \textbf{Baichuan-Omni-1d5}~\cite{li2025baichuan} is another comprehensive model capable of processing text, image, audio, and video inputs and generating text and voice output. It decodes the text and audio simultaneously, which is likely to be slightly different literally but semantically the same. While TTS models and those providing the audio transcription have identical contents in audio and text.

9) \textbf{gpt-4o-audio-preview}~\cite{hurst2024gpt} is a version of GPT-4o with expanded support for audio inputs and the ability to generate text and audio responses.

\begin{table}[h]
    \centering 
    \begin{tabular}{l|r}
    \hline
        \textbf{Topic} & \# \\
    \hline
        Personal Life / Feelings & 821 \\
        Career/Work & 448 \\
        Relationship & 205 \\
        Educational / Study / Life Skill & 174 \\
        Entertainment / Art & 89 \\
        Others & 148 \\
         
    \hline
        \textbf{sum} &  1,885 \\
    \hline
    \end{tabular}
    \caption{Statistics of \ourbench. For \meld and \giga, each audio has three and two different contexts, respectively. For \emovdb, each context has four audios with different tones. Each \textit{(audio, context)} pair is considered one data instance, a total of 1,885.}
    \label{tab:bench-type-topic}
\end{table}

\begin{table*}[!t]
  \centering
  \small
  \begin{tabular}{llccl}
    \toprule
    \textbf{Benchmark} & \textbf{Task} & \textbf{In} & \textbf{Out} & \textbf{Size} \\

    \midrule

    GigaSpeech~\cite{chen2021gigaspeech}      & ASR (En, multi-domain) & A & T &  $\sim$40k h (10k h labeled) \\
    MELD~\cite{poria-etal-2019-meld} & Emotion recognition (multi-party) & A+V+T & T & $\sim$1.4k dialogs; 13k utt. \\
    Emov-DB~\cite{adigwe2018emotional}           & Emotional TTS (controllable) & T & A & 4 speakers; 5 emotions \\
    
    \midrule
    \textbf{\ourbench (Ours)} & \textbf{Empathy response generation} & \textbf{A+T} & \textbf{A/T} & \textbf{1{,}885 instances} \\
    \bottomrule
  \end{tabular}
  \caption{Multi-modality benchmarks. A: Audio; V: Video; T: Text.}
    \label{tab:benchmarks}
\end{table*}

\begin{table*}[t]
    \centering 

    \begin{tabular}{l|cc}
    \hline
       \textbf{Dataset}  &  \textbf{Original Modality}  & \textbf{Contextual Depth}  \\
       \hline
       \meld  & Video with audio, Text & High: multi-turn, multi-party conversations  \\
       \giga & Audio, Text (Transcription)  & Low to Medium (domain focus) \\
       \emovdb & Audio,  Text  & {Low (explicit emotions without context)}  \\
      
    \hline
    \end{tabular}
    \caption{Benchmark features. \emovdb and \meld primarily contain audios with clear emotions.}
    \label{tab:dataset-features-apx}
\end{table*}

\begin{table*}[!h]
  \centering
  \begin{tabular}{clllcccrc}
    \hline
    \textbf{NO.} & \textbf{Model} & \textbf{Input} & \textbf{Output} & \textbf{ASR} & \textbf{Multiturn} & \textbf{Ins.Fo.} & \textbf{Size(B)} & \textbf{Judge} \\
    \hline
    1 & Qwen3-Omni & A, T, I, V & A, T & \Checkmark & \Checkmark & \Checkmark &  35 & \Checkmark\\ %
    
    2 & Qwen2.5-Omni & A, T, I, V & A, T & \Checkmark & \Checkmark & \Checkmark &  10.7 & \Checkmark\\ 
    
    3 & Qwen2-Audio & A, T & T & \Checkmark & \Checkmark & \Checkmark & 8.4 & \Checkmark \\
    
    4 & Qwen-Audio & A, T & T & \Checkmark & \Checkmark & \Checkmark & 8.4 & $\times$ \\

    5 & SALMONN & A, T & T & \Checkmark & $\times$ & \Checkmark &  \textasciitilde 7.0 & $\times$ \\

    6 & Flamingo & A, T & T, TTS & \Checkmark & \Checkmark & \Checkmark & \textasciitilde 7.0 & $\times$ \\

    7 & LLaMA-Omni  & A, T & T, TTS  & $\times$  & \Checkmark  & $\times$  & 9.0 & $\times$ \\

    8 & Baichuan-Omni  & A, T, I, V & A, T & \Checkmark & \Checkmark & \Checkmark & 11.0 & $\times$ \\
    
    \hline
    
    9 & GPT & A, T & A, T & \Checkmark & \Checkmark & \Checkmark & - & \Checkmark \\

    \hline
  \end{tabular}

  \caption{For Input and Output modalities, there are Audio(A, TTS), Text(T), Image(I), and Video(V). For model capabilities, ASR, Multiturn, and Ins.Fo. stand for automatic speech recognition, multi-turn diglogues, and text instruction following, respectively. The model sizes come from HuggingFace (\textasciitilde $ $ denotes approximation). The model versions are Qwen3-Omni-30B-A3B-Instruct, Qwen2.5-Omni-7B, Qwen2-Audio-7B-Instruct, Qwen-audio-chat, SALMONN-7B, LLaMA-Omni2-7B, audio-flamingo-3, Baichuan-Omni-1d5, and gpt-4o-audio-preview, in order. The Judge column indicates whether they are evaluated as a judge model. }
  \label{tab:baseline-comparison}
\end{table*}

\begin{table*}[t]
  \centering
  \setlength{\tabcolsep}{2.5pt}
  \begin{tabular}
  {l|ccccc|ccccc|ccccc}
    \hline
   \multirow{2}{*}{\textbf{Models}} & \multicolumn{5}{c|}{\meld}  & \multicolumn{5}{c|}{\giga} & \multicolumn{5}{c}{\emovdb} \\
   
    &  GPT & Q3-O & Q-O & Q-A &  H & GPT & Q3-O & Q-O & Q-A & H & GPT & Q3-O & Q-O & Q-A & H \\
    \hline

     GPT & \textbf{1.71} & \textbf{1.01} & \textbf{1.09} &  2.72 &    1.46 
     & \textbf{1.88}  & \textbf{1.05} &  \textbf{1.14} &  2.87   & 1.38 
     & \textbf{1.83}  & \textbf{1.00} &  1.09  &  2.83  & 1.90 \\
     
    Qwen3-O & 1.89	& 1.31	& 1.37	& 1.11	& \textbf{1.13}	
    & 1.94	& 1.11	& \textbf{1.14}	& 1.06	& \textbf{1.31}	
    & 1.88	& 1.07	& \textbf{1.08}	& 1.11	& \textbf{1.28} \\
    
     Qwen2.5-O & 1.79 & 1.10 & 1.16  & 2.72  & 1.15 
     & 1.90 & 1.05 &  1.24 &  2.83 & \textbf{1.31} 
     & 1.86 & 1.07 & 1.11  &  2.81  & 1.41 \\ 

\hline

     LLaMA  & 1.92 & 1.32 & 1.45  & 2.74   & 1.67 
     & 1.98 & 1.32 &  1.51 & 2.83  &    1.97  
     & 1.95 & 1.28 &  1.37  &  2.88  &    2.32 \\

     Baichuan  & 2.02  & 1.53 & 1.57  &  2.71  &    1.73 
     & 1.98  & 1.53 &  1.67  &  2.85  & 1.78 
     & 1.99  & 1.48 &  1.43 & 2.84  &  1.95 \\

    \hline
    

    \hline
  \end{tabular}

  \caption{Evaluation on paralinguistic features: \textit{Delivery}. The columns show the judges evaluating audio from models in rows. Judges include GPT-4o-audio-preview (GPT), Qwen3-Omni (Q3-O), Qwen2.5-Omni (Q-O), Qwen2-Audio (Q-A), and Human (H). The level ranges are 1: Good, 2: Fair, 3: Poor. The smaller the value, the better the delivery. Q3-O rates GPT 1.04545 and Qwen2.5-Omni 1.0512, respectively.}
  
  \label{tab:paralinguistic-appendix}
\end{table*}

\begin{table*}[t]
  \centering
  \setlength{\tabcolsep}{3pt}
  \begin{tabular}{l|cccccc|cccccc}
    \hline
\multirow{2}{*}{\textbf{Model}} & \multicolumn{2}{c}{\meld}  & \multicolumn{2}{c}{\giga} & \multicolumn{2}{c|}{\emovdb} &  \multicolumn{2}{c}{\meld}  & \multicolumn{2}{c}{\giga} & \multicolumn{2}{c}{\emovdb} \\
& Coh. & Sup. & Coh. & Sup. & Coh. & Sup. & Coh. & Sup. & Coh. & Sup. & Coh. & Sup. \\
    \hline

GPT & \textbf{1.01} & \textbf{1.01}
& \textbf{1.02} & \textbf{1.00}
& \textbf{1.01} & \textbf{1.00}

& \textbf{1.00} & \textbf{1.00} & \textbf{0.99} & \textbf{1.00} & \textbf{1.00} & \textbf{1.00}\\

Qwen2.5-Omni & 1.07 & 1.03
& 1.03 & 1.03
& 1.02 & 1.02 

& 0.97	& 0.99	& 0.99	& 0.99	& 0.99	& 0.99 \\

Qwen3-Omni & 1.38	& 1.28	& 1.29	& 1.03	& 1.38 & 1.05 
& 0.81	& 0.86	& 0.86	& 0.99	& 0.81	& 0.97\\

LLaMA-Omni  & 1.12 & 1.15
& 1.17 & 1.16 
& 1.12 & 1.13 
& 0.94	& 0.93	& 0.92	& 0.92	& 0.94	& 0.94\\

Baichuan-Omni  & 1.07 & 1.27
& 1.11 & 1.34
& 1.07 & 1.25 
& 0.97	& 0.87	& 0.95	& 0.83	& 0.97	& 0.88\\

    \hline
  \end{tabular}

  \caption{The average scores of Coherence (Coh.) by GPT and Supportiveness (Sup.) by Qwen2.5-Omni. The level ranges are 1: Good, 2: Fair, 3: Poor. The smaller the value, the better. The right part is the normalised scores.}
  \label{tab:coh-sup}
\end{table*}

\section{Evaluation Results}
\label{sec:appendix:good-fair-poor}
The results of Delivery are in Table~\ref{tab:paralinguistic-appendix}. 
The results of Coherence and Supportiveness are in Table~\ref{tab:coh-sup}.

\section{Explanation for Figure~\ref{fig:gigaspeech}}
\label{sec:appendix:figure}
In this example:

1. \textbf{Correctly infer the user’s perspective and intention.}
\begin{itemize}
    \item (text) Go to the store in person.
    \item (text) Get investment info.
\end{itemize}

2. \textbf{Interpret the underlying concerns.}
\begin{itemize}
    \item (audio) people over a certain age → potential wheelchair accessibility concerns.
    \item (audio) more than an acre of → scale of commercial usage.
\end{itemize}

3. \textbf{Adapt the response to be appropriate, relevant, and supportive to address the user’s actual concern.}
\begin{itemize}
    \item (response) Special accessibility support.
    \item (response) Invest in similar properties that occupy a large area.
\end{itemize}

\section{Difference from Instruction Following}
\label{sec:appendix:instruction}

1. \textbf{Cognitive empathy} v.s. \textbf{instruction following}

Instruction following presumes an explicit directive from the user, e.g., “Provide…,” “Tell me…,” “Answer…,” “Give me…,” with a clearly defined task-oriented goal.

By contrast, cognitive empathy involves the ability to:
\begin{itemize}
    \item infer the user’s unstated goal, concern, or perspective,
    \item interpret ambiguous or underspecified context, and
    \item respond in a way that aligns with the user’s internal state, even when it is not explicitly articulated.
\end{itemize}

2. \textbf{Why this distinction applies to AEQ-Bench}

In our setting, the context is a compressed summary of prior dialogue, but it contains no explicit instructions. The model must therefore infer the user’s intention from clues in the context + utterance. In example (Fig.~\ref{fig:gigaspeech}), the model pays attention to different parts of the same utterance according to different contexts.

The model should read between the lines to infer these concerns when the user does not instruct “Please check accessibility” or “Please provide investment details”.

Thus, the task is not a direct instruction-following exercise but a test of whether the model can: i) understand the user’s perspective, 
ii) detect latent concerns, and
iii) craft a contextually supportive reply.

3. \textbf{Relation to our broader goal}

Most real human conversations are not phrased as tasks or explicit instructions.

People frequently imply their concerns, leave intentions unstated, or expect the listener to infer what they need. For example, when someone says, “That café is up a really steep hill” they rarely add request or instruction, “Please check accessibility for me.” A supportive interlocutor is expected to read between the lines and recognize the underlying concern. This ability—inferring perspective, intention, and concern from context without an explicit directive—is precisely what we aim to evaluate through the GigaSpeech subset.

AEQ-Bench is designed to evaluate EQ-oriented companion AI, which requires sensitivity to user perspectives, goals, and concerns/worries, even when emotions are implicit. This complements the prevailing focus on Instruction-following and IQ-oriented logic and reasoning evaluations.

\section{Construction Steps}
\label{sec:appendix:construction}
\paragraph{Audio filtering.}
We only retain audios with clear delivery to prevent adding unnecessary difficulty to the speech recognition stage.
We exclude all instances of the sleepiness tone present in the \emovdb dataset, as these are typically slow, slurred, and often contain non-speech acoustic cues (e.g., yawns).

On the other hand, data instances that require high-level reasoning or factual complexity were excluded. 
Specifically, :  

(i) Asking the other party to perform physical actions. (e.g., \textit{pick up my friends.})
 
(ii) Mathematical reasoning or complex task completion. (e.g., \textit{how to arrange the chores to finish them efficiently}; \textit{booking a flight.})
 
(iii) Target at real persons, such as intimate confessions, family disputes. ( e.g., \textit{You are the most charming woman in the world.}) 

\meld and \emovdb are filtered manually, and \giga is filtered by GPT in early rounds. 

\paragraph{Quality Validation Annotators.}

Three in-house annotators, expertising in linguistics, conducted the verification.
They were compensated at 16 USD/hour, the same rate as response evaluation annotators.
Approximately 70\% of the automatically generated contexts were judged coherent. The remaining contexts were manually revised by the annotators to ensure quality before being used for evaluation.

\section{Details of Evaluation Metrics}
\label{sec:appendix:metric}
Empathy is inherently multi-dimensional. Therefore, it naturally requires both categorical decisions and graded judgments:

\paragraph{Categorical decisions} reflect discrete states (i.e., which modality the model is relying on). Specifically, Modality Reliance: Did the model rely on the audio or the text? We need to know which modality OLMs utilize to give a response in conversation, since previous studies only include instructions in the other modality, leaving it unclear whether context information is involved.

\paragraph{Graded qualities} capture nuance in empathic responding that cannot be captured by simple classifications (e.g., how supportive, how natural, how context-adaptive). Ordinal scales allow for subtle differentiation in degree (e.g., “good,” “fair,” “poor”; or numeric gradations) that reflect human judgments of how well a response makes them feel, not just whether it does or doesn’t. (e.g., Barrett-Lennard Relationship Inventory~\cite{chen2023development}, Mean Opinion Score (MOS)~\cite{viswanathan2005measuring}.

Most rating metrics in AEQ-Bench (i.e., Coherence, Supportiveness, Delivery) use three levels (Good, Fair, Poor) because these qualities tend to vary along coarse but meaningful interpersonal distinctions. This mirrors standard practice in social psychology and counseling assessment tools, where empathy-related constructs are commonly evaluated using 3-point or 5-point Likert scales to capture the essential gradation without overburdening annotators.

For example, Naturalness reflects a more nuanced, continuous progression in conversational behavior, from pure analytical, to analytical+Chatting, to AI-like Chatting, to human-like Chatting. Annotators consistently reported that these degrees of difference are clear and distinguishable, warranting a finer-grained ordinal scale.

\paragraph{Why these metrics are chosen? }

\paragraph{Modality Reliance.} Recognizing and using both vocal and textual cues ensures the system is sensitive to how emotion is expressed.
\paragraph{Coherence.} essential for empathetic accuracy~\cite{ickes2000difficulty}. If a response is incoherent to the context, it likely fails to reflect empathic understanding.

\paragraph{Discrimination.} Empathic accuracy is not just about detecting emotion, but differentiating subtle differences depending on context~\cite{ickes2000difficulty}. A model must adapt to context shifts (or changes in user tone or emotion) rather than produce generic responses.

\paragraph{Naturalness.} More human-like language or voice creates a stronger sense of connection and perceived empathy~\cite{kuhne2020human}.

\paragraph{Supportiveness.} Counseling psychology defines empathy in part as validation + perspective-taking + nonjudgmental support~\cite{rogers1957necessary}.

\paragraph{Delivery and Emotion}: Paralinguistic features like tone, pause, and pitch significantly influence perceived empathy in human speech~\cite{burleson2009revisiting}. Emotional expressiveness, rather than neutral tone, plays a crucial role in how people perceive empathy~\cite{loveys2021effects}.

\section{Evaluation Metric Scores}
\label{sec:appendix:scores}

This appendix provides a detailed breakdown of the evaluation metrics used in our human and automated assessments, including the specific questions posed to annotators, the corresponding scale or options, and the precise criteria for each score. The metrics are divided into linguistic features (focusing on the content of the response) and paralinguistic features (focusing on the acoustic delivery).

\subsection{Linguistic Features}

\paragraph{\textbf{Q1. Modality Reliance (M.R.)}} This metric evaluates which input source the model primarily utilizes to generate its textual response, verifying if it relies on the provided conversation context, the user's immediate utterance (ASR transcription), or a combination of both.

\noindent \textbf{Evaluation Question}: Which input modality (Context or Utterance) does the model primarily rely on, or does it utilize both, to generate the response?

\noindent \textbf{Scale and Criteria}:
\begin{enumerate}
\item \textbf{Context}: Response is based only on the provided CONTEXT.
\item \textbf{Utterance}: Response is based only on the current UTTERANCE (ASR).
\item \textbf{Both}: Uses CONTEXT and UTTERANCE together, showing no obvious conflict.
\item \textbf{Failed}: Response is irrelevant, off-topic, a repetition/translation of inputs, or includes disclaimers.
\end{enumerate}

\vspace{0.8em} 

\paragraph{\textbf{Q2. Naturalness (Nat.)}} Naturalness assesses how human-like and spontaneous the generated response is, simulating a natural conversation between peers. Higher scores indicate a more human-like exchange.

\noindent \textbf{Evaluation Question}: How natural and human-like is the response? The more human-like the conversation, the higher the score.

\noindent \textbf{Scale and Criteria}: NA, 1 to 4.
\begin{itemize}
\item \textbf{NA (Not Applicable)}: The response is irrelevant or a repetition/translation of the inputs.
\end{itemize}
\begin{enumerate}
\item \textbf{Score 1 (Analysis only)}: The response is pure comment or analysis, with no direct conversational reply.
\item \textbf{Score 2 (Analysis + Chatting)}: A mix of third-person analysis and first-person reply.
\item \textbf{Score 3 (AI-like Chatting)}: Direct reply with clear, formulaic Artificial Intelligence (AI) characteristics.
\item \textbf{Score 4 (Human-like Chatting)}: High naturalness, closely resembling a fluent human conversation.
\end{enumerate}

\vspace{0.8em}

\paragraph{\textbf{Q3. Coherence (Coh.)}} Coherence judges the logical and emotional consistency of the response's content relative to the entire conversation and the user's current utterance.

\noindent \textbf{Evaluation Question}: Is the response logically and emotionally consistent with the overall CONTEXT and UTTERANCE?

\noindent \textbf{Scale and Criteria}: NA, Good, Fair, Poor.
\begin{itemize}
\item \textbf{NA (Not Applicable)}: The response is irrelevant or a repetition/translation of the inputs.
\end{itemize}
\begin{enumerate}

\item \textbf{Good}: Clearly relevant, logically consistent, and appropriate to the situation.
\item \textbf{Fair}: Mostly relevant but with minor inconsistencies or vague links to the context.
\item \textbf{Poor}: Off-topic, contradictory, or highly inappropriate given the inputs.
\end{enumerate}

\vspace{0.8em}

\paragraph{\textbf{Q4. Supportiveness (Sup.)}} This metric evaluates the degree of supportive intent conveyed by the response's text, considering factors such as emotional acknowledgment, perspective-taking, supportive intent (comfort/encouragement/help), and non-judgmental language.

\noindent \textbf{Evaluation Question}: How supportive does the response feel? Considering acknowledgment of feelings, perspective-taking, supportive intent, and non-judgmental language.

\noindent \textbf{Scale and Criteria}: NA, Good, Fair, Poor.
\begin{itemize}
\item \textbf{NA (Not Applicable)}: The response is irrelevant or a repetition/translation of the inputs.
\end{itemize}
\begin{enumerate}

\item \textbf{Good}: Clear acknowledgment, perspective-taking, strong supportive intent, and non-judgmental tone.
\item \textbf{Fair}: Some supportive elements, but incomplete, formulaic, or mixed with minor judgment.
\item \textbf{Poor}: Lacks support, is dismissive, minimizing, or overtly judgmental.
\end{enumerate}

\vspace{0.8em}

\paragraph{\textbf{Q5. Discrimination (Disc.)}} Discrimination measures the model's ability to vary its textual content based on differing contexts or emotional cues in the user's input. A high score indicates the response is highly tailored.

\noindent \textbf{Evaluation Question}: Across varying contexts/tones, how much does the model's response content vary? The more tailored for context/tone, the higher the score.

\noindent \textbf{Scale and Criteria}: NA, 1 to 6.
\begin{itemize}
\item \textbf{NA (Not Applicable)}: Response is irrelevant/repetition.
\end{itemize}
\begin{enumerate}
\item \textbf{Score 1 (No difference)}: Content remains identical across inputs.
\item \textbf{Score 2 (Minor wording)}: Same core content, only small lexical changes.
\item \textbf{Score 3 (Some variations)}: Mostly same content, but with isolated tailored variations.
\item \textbf{Score 4 (Mostly different)}: Responses show significant differences, adapting to context.
\item \textbf{Score 5 (Template-like)}: Clear differences, but the wording is highly standardized or template-based.
\item \textbf{Score 6 (Contextually adapted)}: Clear, diverse differences with contextually and emotionally appropriate, varied wording.
\end{enumerate}

\vspace{0.8em}

\subsection{Paralinguistic Features (Audio Assessment)}





\paragraph{\textbf{Q6. Delivery}} Delivery assesses the acoustic quality and appropriateness of the speech synthesis, focusing on how paralinguistic cues (tone, pacing, prosody, and timing) contribute to the overall supportive or emotional intent of the response.

\noindent \textbf{Evaluation Question}: How supportive does the response \textit{sound}? Does the tone of voice, pacing, pitch, and pauses effectively convey empathy or the intended emotion?

\noindent \textbf{Scale and Criteria}: NA, Good, Fair, Poor.
\begin{itemize}
\item \textbf{NA (Not Applicable)}: The response is irrelevant or a repetition/translation of the inputs.
\end{itemize}
\begin{enumerate}

\item \textbf{Good}: Warm, calm tone; patient pacing; appropriate prosody; natural pauses and timing.
\item \textbf{Fair}: Generally acceptable but inconsistent (e.g., slightly rushed/flat tone or occasional awkward pauses).
\item \textbf{Poor}: Cold/harsh tone, stilted or mechanical pacing, mismatched prosody, or unnatural timing.
\end{enumerate}

\section{Human Annotation for Responses}
\label{sec:appendix:annotation}
Specifically, the five OLMs are grouped in pairs, resulting in 10 model pairs. 
We sample two groups of data for each mdoel pair. Each grouped data consists of one utterance associated with three contexts for \meld, two for \giga, and four tonal variations plus one context for \emovdb. A total of 180 instances and responses. 

Based on the provided context, utterance, response audio, and evaluation instructions, the annotators assess the responses across all metrics. All annotators are fluent English speakers recruited from university student helpers; a total of 12 annotators participated.

Cohen’s Kappa (for categorical metrics): modality\_dependency=0.72. 

ICC (for ordinal scores): naturalness = 0.84, coherence=0.89, supportiveness=0.87, delivery=0.86, discrimination=0.80.

OLM judges are instructed to evaluate the responses with the same set of instructions as human annotators'.

\section{Fine-grained Paralinguistic Evaluation}
\label{sec:appendix:fine-grained}
This experiment examine whether emotional and paralinguistic qualities of synthesized speech can be faithfully evaluated through textual intermediates, rather than direct auditory perception.

Specifically, we compare: (i) human listening as the gold standard; (ii) direct audio judging by an OLM; (iii) free-form paralinguistic descriptions generated by an OLM; and (iv) constrained, objective textual descriptions of acoustic properties. For each setting, we compute 5-point paralinguistic quality scores across MELD, GigaSpeech, and EmoV-DB subsets.
5-point scale in Table~\ref{tab:para-scale}. 

\begin{table}[h]
\centering \small
\begin{tabular}{c|p{1cm}|p{5cm}}
\hline
\multicolumn{2}{l|}{\textbf{Score}}  & \textbf{Evaluation} \\
\hline
5 & Very Good & Paralinguistic cues are highly expressive, perfectly conveying emotion and intent, resulting in smooth and natural communication (human-like). \\
4 & Good & Paralinguistic cues are mostly clear. Slightly more emotional than typical machine speech, but still not fully natural. \\
3 & Fair & Paralinguistic cues are basically conveyed and do not hinder understanding (typical machine-like speech). \\
2 & Poor & Paralinguistic cues are vague and frequently interfere with communication, making understanding difficult. \\
1 & Very Poor & Paralinguistic cues are almost unrecognizable and severely impair communication effectiveness. \\
\hline
\end{tabular}
\caption{Five-point scale for paralinguistic evaluation.}
\label{tab:para-scale}
\end{table}


\clearpage

\end{document}